%% file: main.tex
\documentclass[10pt,twocolumn,letterpaper]{article}

\usepackage[pagenumbers]{cvpr} 

\input{preamble}

\definecolor{cvprblue}{rgb}{0.21,0.49,0.74}
\usepackage{pifont}
\usepackage{float}
\usepackage{paralist}
\usepackage{amsmath}
\usepackage{amssymb}
\usepackage{algorithm}
\usepackage{algorithmicx}
\usepackage{algpseudocode}
\usepackage{booktabs}
\usepackage{multicol}
\usepackage{multirow}
\usepackage{makecell}
\usepackage{subcaption}
\usepackage{graphicx}
\usepackage{xcolor}
\usepackage{setspace}
\usepackage{listings}
\usepackage{subcaption}

\usepackage[margin=0.75in]{geometry}
\usepackage{times}
\usepackage[utf8]{inputenc}
\usepackage[T1]{fontenc}
\usepackage{graphicx}
\usepackage{amsmath,amsfonts,amssymb}
\usepackage{booktabs}
\usepackage{url}
\usepackage{xcolor}
\usepackage{subcaption}

\newcommand{\Tref}[1]{Table~\ref{#1}}
\newcommand{\Eref}[1]{Eqn.~(\ref{#1})}
\newcommand{\Fref}[1]{Figure~\ref{#1}}
\newcommand{\Sref}[1]{Sec.~\ref{#1}}
\newcommand{\Aref}[1]{Algorithm~\ref{#1}}

\usepackage[colorlinks=true,citecolor=blue,urlcolor=blue]{hyperref}

\title{PointCNN++: Performant Convolution on Native Points}

\author{Lihan Li$^{1,2,\#*}$~~
Haofeng Zhong$^{1,2,\#*}$~~
Rui Bu$^{2}$~~
Mingchao Sun$^{3}$\\
Wenzheng Chen$^{1,\dagger}$~~
Baoquan Chen$^{1,\dagger}$~~
Yangyan Li$^{2,\dagger}$\\
$^1$ Peking University~~$^2$ Ant Group~~$^3$ AMAP\\
{\small \texttt{\{hfzhong, wenzhengchen, baoquan\}@pku.edu.cn, \{burui.br, yangyan.lyy\}@antgroup.com,}}\\
\small \texttt{lh\_li@stu.pku.edu.cn, sun.mc@outlook.com}}

\begin{document}
\maketitle
\input{sec/0_abstract}
{
    \renewcommand*{\thefootnote}{\#}
    \footnotetext{Equal contributions.\,$^{\dagger}$\,Corresponding author.\,$^{*}$\,Work done during internship at Ant Group.}
}
\input{sec/1_intro}
\input{sec/2_related}
\input{sec/3_method_new}
\input{sec/4_exps}
\input{sec/5_conclusion}

\renewcommand{\thesection}{\Alph{section}}
\renewcommand{\thefigure}{\Alph{figure}}
\renewcommand{\thetable}{\Alph{table}}
\renewcommand{\thealgorithm}{\Alph{algorithm}}
{
    \small
    \bibliographystyle{ieeenat_fullname}
    \bibliography{main}
}


\end{document}

%% file: preamble.tex
\newcommand{\red}[1]{{\color{red}#1}}



%% file: sec/0_abstract.tex
\begin{abstract}
Existing convolutional learning methods for 3D point cloud data are divided into two paradigms: point-based methods that preserve geometric precision but often face performance challenges, and voxel-based methods that achieve high efficiency through quantization at the cost of geometric fidelity. This loss of precision is a critical bottleneck for tasks such as point cloud registration. We propose PointCNN++, a novel architectural design that fundamentally mitigates this precision-performance trade-off. It \textbf{generalizes sparse convolution from voxels to points}, treating voxel-based convolution as a specialized, degraded case of our more general point-based convolution.
First, we introduce a point-centric convolution where the receptive field is centered on the original, high-precision point coordinates. Second, to make this high-fidelity operation performant, we design a computational strategy that operates \textbf{natively} on points. We formulate the convolution on native points as a Matrix-Vector Multiplication and Reduction (MVMR) problem, for which we develop a dedicated, highly-optimized GPU kernel.
Experiments demonstrate that PointCNN++ \textbf{uses an order of magnitude less memory and is several times faster} than representative point-based methods. Furthermore, when used as a simple replacement for the voxel-based backbones it generalizes, it \textbf{significantly improves point cloud registration and semantic segmentation accuracies while proving both more memory-efficient and faster}.
PointCNN++ shows that preserving geometric detail and achieving high performance are not mutually exclusive, paving the way for a new class of 3D learning with high fidelity and efficiency.

Our code is publicly available at:~\url{https://github.com/ant-research/pointelligence}
\end{abstract}

%% file: sec/1_intro.tex
\section{Introduction}
\label{sec:intro}

\begin{figure*}
    \centering
    \includegraphics[width=0.9\linewidth]{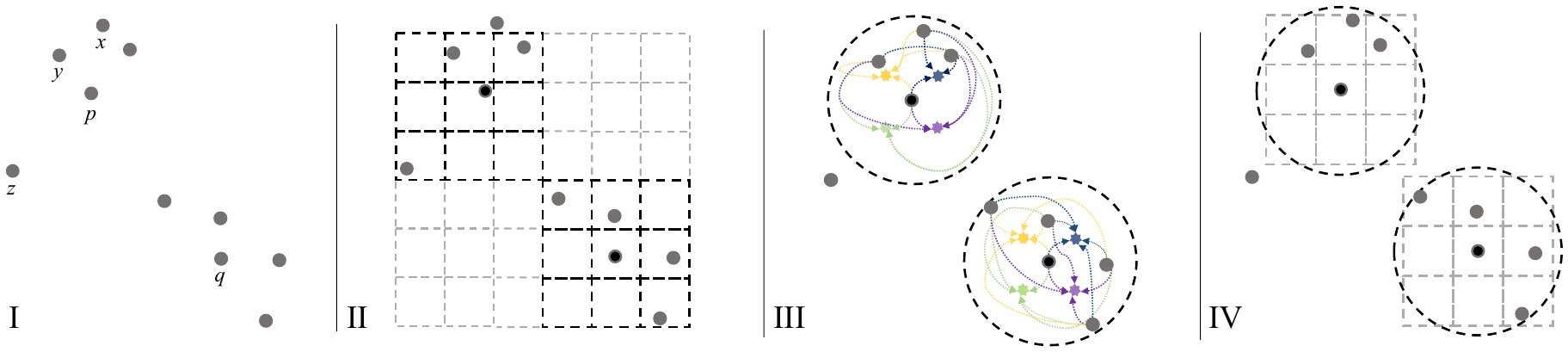}
    \vspace{-4mm}
    \caption{A 2D illustration of convolutional learning for point cloud (I) with voxel-based methods (II), transform-then-convolve methods (III) and convolution on native points (IV). When a voxel center happens to be on an original point (the rare case, as depicted by $q$), the difference between (II) and (IV) is minimal. However, in the general cases, due to the forceful restricting of computation on voxel grids in (II), several problems arise: 1. misalignment between original points (\emph{e.g.}, point $p$) and convolution centers, 2. inaccurate neighborhood inclusion(\emph{e.g.}, $x$, instead of $z$, should be in the neighborhood of $p$), and 3. inaccurate convolution kernel usage (\emph{e.g.}, the feature associated with point $x$ is more appropriate for being convolved with the upper-left kernel as show in (IV), instead of the upper-middle kernel as shown in (II). IV preserves geometric precision as those in III, while avoiding the cumbersome irregular-to-regular ``transformation''.}
    \label{fig:conv}
    \vspace{-5mm}
\end{figure*}

The proliferation of 3D sensing technologies has established point clouds as a primary data modality in domains like autonomous driving~\cite{huang2018apolloscape,guo2022lirtest,yang2024visual}, robotics~\cite{pomerleau2015review,kim2018slam,zhu2024point,lin2025autourdf}, and augmented reality~\cite{placitelli2011low,alexiou2017towards,wang2023pointshopar,peek2025novel}. As a direct representation of world geometry, a point cloud's value is intrinsically tied to its high-fidelity spatial information. This very characteristic---inherently sparse and irregular data---however, creates a fundamental challenge for modern computational hardware~\cite{nvidia_hopper_2022,nvidia_ampere_2020,jouppi2023tpu,amd_cdna2_2021} and software~\cite{nickolls2008scalable,jia2014caffe,abadi2016tensorflow,paszke2019pytorch} that are heavily optimized for dense and regular data.

To address this challenge, the study of convolution on point clouds has developed two competing paradigms, each embodying a significant compromise. The most prevalent, the voxel-based approach (Figure~\ref{fig:conv}  II), resolves the conflict by \textbf{forcefully restricting the convolution on voxel grids}. This approach firstly applies a \emph{global voxelization} to quantize an entire continuous space of interest into a set of sparse voxels, from which sparse convolutional operators are applied. While the performance issue of such convolution has been substantially addressed by leveraging the sparse nature of the data, with representative work from O-CNN~\cite{wang2017cnn}, SPConv~\cite{spconv2022}, to MinkowskiEngine~\cite{choy20194d}, the quantization is an inherently lossy sampling operation, resulting in impaired fine geometric details—representing high-frequency spatial signals. The act of the global voxelization in the beginning of the process establishes an a priori error floor determined by the voxel size, posing a critical bottleneck to tasks like high-precision registration that depend on sub-voxel feature uniqueness.

An alternative philosophy, the point-based paradigm, attempts to preserve the data's integrity by a relatively more gentle, usually learned, transformation of irregular points into a regular dense tensors, from which the convolution is then applied. Such transform-then-convolve approach (Figure~\ref{fig:conv} III) is shown to be effective in harvesting the fidelity in point clouds, as shown in the notable representative work of PointCNN~\cite{li2018pointcnn} and KPConv~\cite{thomas2019kpconv}. However, while the convolution on the dense tensor is indeed computationally efficient, \textbf{significant inefficiency arises in the transformation from irregular to regular itself}. Such transformation introduces extra computation, parameters to learn and significant memory access---a major source of inefficiency in point cloud related computation on GPUs~\cite{lin2021pointacc,tang2022torchsparse}. Moreover, the implementation of such point-based methods often involves frequent usage of padding operations, which often intensifies the inefficiency.

This paper posits that this prevailing trade-off should not be viewed as a permanent compromise, but rather as a conflict that can be mitigated through holistic computational design.
We introduce PointCNN++ (Figure~\ref{fig:conv} IV), an architectural design that resolves this conflict by advancing a new computational paradigm. Instead of \textbf{restricting the convolution} or \textbf{introducing extra transformation}, we design the operator and compute kernel as an integrated system, purpose-built for the high-performance processing of native point clouds, without any degradations nor superfluities. Our approach begins by generalizing sparse convolution from discrete voxels to continuous points, centering operations on the true, high-precision coordinates as much as possible. As the last operation, we use a \emph{local and adaptive voxelization} to pair the data to be convolved with the discrete convolution kernels is used as the last operation to minimize the fidelity loss. From this perspective, voxel-based convolution is merely a quantized, degraded special case of our more fundamental point-centric operator. A more general operator, however, is only practical if it is performant. To this end, we formulate the convolution on native points as a Matrix-Vector Multiplication and Reduction (MVMR) problem. With this formulation, we drew inspiration from an efficient algorithm~\cite{RIVERA202170} for the Matrix-Vector Multiplication sub-problem to develop a dedicated, highly-optimized GPU kernel for the full problem.

In this paper, we introduce PointCNN++, an architectural design that mitigates the long-standing conflict between precision and performance in convolutional learning for 3D point cloud data. 
We empirically demonstrate that PointCNN++ not only excels on precision-critical tasks such as registration but also delivers state-of-the-art results on large-scale semantic segmentation benchmarks. Moreover, it is more memory-efficient while being even faster than existing approaches.
This work proves that by holistically designing the computational system of point cloud data, achieving high geometric fidelity and high performance is not a mutually exclusive goal. 
\vspace{-3.5mm}

%% file: sec/2_related.tex
\section{Related Work}
\subsection{Feature Learning for 3D Point Cloud}
Deep learning on point clouds has evolved through several operator paradigms.
The seminal PointNet~\cite{qi2017pointnet} processes each point independently with shared MLPs before global aggregation, while its successor PointNet++~\cite{qi2017pointnet++} introduces a hierarchical structure by recursively applying this process on nested point subsets.
A central line of work focuses on generalizing convolution operator for 2D image data into processing 3D data.
Voxel-based methods first discretize space for efficient 3D CNNs; for instance, VoxelNet~\cite{zhou2018voxelnet} learns a unified feature representation for points within each voxel.
PointCNN~\cite{li2018pointcnn} proposes a learned $\mathcal{X}$-Transform to permute local point features into a canonical order before applying a convolution. PointConv~\cite{wu2019pointconv} learns continuous kernel weights from relative coordinates using an MLP. KPConv~\cite{thomas2019kpconv} uses a set of rigid, learnable kernel points to apply spatially-aware weights.
DGCNN~\cite{wang2019dynamic} constructs dynamic graphs in feature space, and applies convolution on the feature space neighborhoods.

While Transformer has been widely used in processing 3D point cloud data, one thing to note is that they often operate on the features extracted with convolutional backbones~\cite{zhao2021point,huang2021predator,zhang2021point,Zhou_centerformer,lai2022stratified,9861747,liu2023flatformerflattenedwindowattention}, without which their effectiveness has not been widely demonstrated.
Our work is most closely related to convolutional backbones in general, either used alone or as a part of a larger architecture, for serving general feature learning purposes on point cloud data. We show that with our systematic design, the performance advantage of the voxel-based methods and the precision advantage of point-based methods could be combined.

\vspace{-1mm}
\subsection{Performant Computation on Sparse Voxels}
The performance of 3D deep learning is critically dependent on their underlying computational systems. The sparse nature of 3D data is extensively explored to efficiently process data while avoiding computation on empty space.
SparseConvNet~\cite{graham2017submanifold} pioneered the work in this domain, introducing the use of hashmaps to manage the coordinates of active voxels.
Building on this, SpConv~\cite{yan2018second} proposed a highly-optimized grid-based map search and formalized the influential gather-matmul-scatter dataflow.
MinkowskiEngine~\cite{choy20194d} significantly improved upon SparseConvNet's hashmap implementation to reduce latency and introduced an alternative fetch-on-demand dataflow.
More recent frameworks like TorchSparse~\cite{tang2022torchsparse} and TorchSparse++~\cite{tangandyang2023torchsparse} have continued to push performance boundaries by developing novel traversal algorithms, hand-tuned CUDA kernels, and co-optimizing data structures and workloads specifically for GPU architectures.
$f$VDB~\cite{williams2024fvdb} integrated these concepts and further developed a flexible framework accepting various popular 3D data representations into the unified voxel-based representation upon which a rich set of operators is based.
While existing high-performance methods typically exploit sparsity via voxelization, we show, however, that this is not a necessary coupling. Our approach achieves comparable computational efficiencies by leveraging sparsity directly on native point cloud representations, thereby bypassing the geometric precision loss inherent to voxelization.
\vspace{-2mm}

%% file: sec/3_method_new.tex
\section{Method}
\label{sec:method}
This section details the design principles of PointCNN++. Our methodology resolves the long-standing precision-performance dichotomy in 3D deep learning through a synergistic co-design of the core operator and its underlying computational system. We begin by establishing a generalized view of convolution and situating prior work within this framework in~\Sref{sec:preliminaries}. We then introduce our point-centric convolution, highlighting its properties as a powerful generalization in~\Sref{sec:point_centric_conv}. Finally, we describe the performant system design in~\Sref{sec:system_design}, from the data representation to the optimized GPU kernel, that makes this high-fidelity operator computationally feasible at scale.

\subsection{Preliminaries on Convolution}
\label{sec:preliminaries}
\subsubsection{A Generalized View of Convolution}
At its core, a convolution operator computes a feature $\mathbf{F}^{\text{out}}_i \in \mathbb{R}^{C_{\text{out}}}$ at location $\mathbf{P}^{\text{out}}_i$ by aggregating information from neighborhood locations $\{\mathbf{P}^{\text{in}}_j\}$, each associated with a feature $\mathbf{F}^{\text{in}}_j \in \mathbb{R}^{C_{\text{in}}}$, with the help of some learnable kernels $\{\mathbf{W}_k \in \mathbb{R}^{C_{\text{out}} \times C_{\text{in}}}\}$. $C_{\text{in}}$ and $C_{\text{out}}$ are dimensions, or channels, of the input and output features, respectively. The computational interactions among $\mathbf{F}^{\text{out}}_i$, $\mathbf{F}^{\text{in}}_j$, and $\mathbf{W}_k$, could be completely depicted by a triplet set $\mathcal{T} = \{(i, j, k)\}$, where a triplet $(i, j, k)$ denotes that for convolution at location $\mathbf{P}^{\text{out}}_i$, location $\mathbf{P}^{\text{in}}_j$ is in its neighborhood, and $\mathbf{W}_k$ is the corresponding kernel to use for this neighborhood location. Thus, a generalized view of convolution could be formulates as:
\vspace{-2mm}
\begin{equation}
\label{eq:generalized_conv}
\mathbf{F}^{\text{out}}_i = \sum_{(i, j, k) \in \mathcal{T}} \mathbf{W}_k \times \mathbf{F}^{\text{in}}_j.
\end{equation}
The essence of different convolution types lies in how the triplet set $\mathcal{T} = \{(i, j, k)\}$ is constructed. More specifically, there are three major considerations in the construction of such triplets:~\ding{182} How the locations of output features are placed? --- Considerations on defining convolution centers $\{\mathbf{P}^{\text{out}}_i\}$ to use.~\ding{183} How are the neighborhood measured? --- Considerations on the pairing of $\{(i,j)\}$ in the triplet.~\ding{184} Which kernel to use for a location in the neighborhood? --- Considerations on the pairing of $\{(j,k)\}$ in the triplet.

\subsubsection{Convolution on 2D Images}
Convolution on 2D images is a specialized instance of~\Eref{eq:generalized_conv}, where the triplet set $\{(i, j, k)\}$ is constructed by taking fully considerations on the dense and regular nature of the image data that is ubiquitously represented as discrete pixels:~\ding{182} $\{\mathbf{P}^{\text{out}}_i\}$ are the pixels.~\ding{183} $\{(i,j)\}$ are predefined as the pixels in patches around $\{\mathbf{P}^{\text{out}}_i\}$ are taken as neighborhoods,~\emph{i.e.}, Chebyshev distance based neighbors.~\ding{184} For $\{(j,k)\}$, the kernels and neighborhoods are often of the same spatial resolution, thus $\{\mathbf{W}_k\}$ are paired with $\{\mathbf{P}^{\text{in}}_j\}$ if they share the same spatial locations.

Significant efforts have been made to achieve efficient GPU algorithms for the specialized triplet set $\mathcal{T}$ of image convolution. In the early development stage of deep learning, image convolution was~\emph{not} a highly optimized standard operation on GPU. One way to achieve efficient computation was to materialize $\mathbf{F}^{\text{in}}_j$, thus lower the convolution into an highly optimized general matrix-matrix multiplication (GEMM), as practiced in the early version of Caffe~\cite{jia2014caffe}. However, as detailed in cuDNN~\cite{chetlur2014cudnn}, such materialization inevitably introduces significant amount of extra GPU memory usage, which could be addressed by lazily materializing in on-chip memory only.
\vspace{-1mm}
\subsubsection{Convolution on 3D Sparse Voxels}
With a \emph{global voxelization} of the entire 3D space of interest, often in the form of point cloud, the space could be sampled into sparse voxels --- a representation that is a generalization of 2D pixels into 3D. By adding one dimension to the spatial coordinates, and relaxing the assumption of dense and regular tensor, convolution operator can be generalized from 2D images into 3D sparse voxels. Yet, convolution on 3D sparse voxels is still a specialized instance of~\Eref{eq:generalized_conv}, as illustrated in~\Fref{fig:conv} II:~\ding{182} $\{\mathbf{P}^{\text{out}}_i\}$ are the discrete voxels.~\ding{183} $\{(i,j)\}$ are constructed by taking the non-empty voxels around $\{\mathbf{P}^{\text{out}}_i\}$ as neighborhoods,~\emph{i.e.}, again, Chebyshev distance based neighbors.~\ding{184} Same as that in convolution on 2D images, $\{(j,k)\}$ are constructed by corresponding same spatial locations.

While efficient GPU algorithms of voxel-based convolution have been proposed in notable representative work from O-CNN~\cite{wang2017cnn}, SPConv~\cite{spconv2022}, to MinkowskiEngine~\cite{choy2019fully,choy2020high}. There are inherent drawbacks arise from the definition of voxel-based convolution: (1) placing $\{\mathbf{P}^{\text{out}}_i\}$ at the discrete voxels is at the cost of sacrificing fidelity from the original point cloud; (2) neighborhood construction in Chebyshev distance with imprecise centers intensifies the fidelity erosion; (3) the fineness of kernels are coupled with the fineness of voxelization. A more detailed discussion, and addressing, of such drawbacks are given after the introduction of our method in the next section.
\vspace{-2mm}
\subsection{Convolution on Native Points}
\label{sec:point_centric_conv}
\noindent\textbf{Definition.} Our design philosophy is to fully harvest the fidelity in the input point cloud. As illustrated in~\Fref{fig:conv} IV, We start by~\ding{182} placing the convolution locations right at the original high-precision points, thus $\{\mathbf{P}^{\text{out}}_i \in \mathbb{R}^3\}$ are true, continuous coordinates in our formulation. Then,~\ding{183} the pairing of $\{(i,j)\}$ could be constructed based on neighborhood search using the precise convolution centers with appropriate distance metrics on continuous space. Finally~\ding{184}, for constructing $\{(j,k)\}$, a \emph{local voxelization}, centered right at $\mathbf{P}^{\text{out}}_i$ of the same spatial resolution as the kernels is applied in each neighborhood region, yielding a correspondence between the kernels $\{\mathbf{W}_k\}$ and $\{\mathbf{P}^{\text{in}}_j\}$ in the neighborhood voxels.

\noindent\textbf{Advantages.} It is clear that both~\ding{182} and~\ding{183} operates on the full precision of the original point cloud, the quality of the neighborhood regions are well preserved. It is only at the final step,~\ding{184}, with the convolution centers align exactly with neighborhood region centers, a \emph{local voxelization} is applied. Such a local voxelization is adaptive to each neighborhood region, resulting in a higher quality mapping between $\{\mathbf{W}_k\}$ and $\{\mathbf{P}^{\text{in}}_j\}$. Note that, with given neighborhood regions, it is the appropriate fineness of the convolution kernels that defines the resolution of the local voxelization, rather than being coupled to the fineness of the global voxelization as that in convolution on sparse voxels.

\noindent\textbf{Voxel Convolution as a Special Case.} With the definition of convolution on native points, now we show that convolution on voxels is a special case of convolution on native points, by demonstrating the three degradations have to take that convert convolution on native points into convolution on voxels. First of all, such conversion inevitably introduce a global voxelization to generate voxels that are not necessary for convolution on the native points, but mandatory for convolution on voxels. Degradation~\ding{182}, instead of using the original points as the convolution centers, snapping the convolution centers into the centers of the voxels. Degradation~\ding{183}, instead of searching neighbors with the original points as centers within other points, searching neighbors in the voxel space. Degradation~\ding{184}, instead of choosing appropriate fineness thus resolution of the kernels, using the resolution that is coupled to the size of receptive field in the global voxelized space.

For degradation~\ding{184}, in other words, convolution on native points decouples kernel resolution from the receptive field's physical span, whereas convolution on voxels couples it to the voxel resolution in the receptive field. We demonstrate the difference with some  examples. In convolution on native points, given a neighborhood of points, kernels with fineness of either $3^3$ or $5^3$ could be used up to demand, as the continuous coordinates could result in any of the local voxels. In contrast, in convolution on voxels, if a neighborhood is constructed with $3^3$ voxels based on the global voxelization, it does not make sense to use kernels of $5^3$ resolution to convolve with such a neighborhood, as the neighborhood has been quantized into the coarser $3^3$ voxels, leaving the extra fineness of $5^3$ kernels useless.
\vspace{-2mm}
\subsection{Performant Systematic Design}
\label{sec:system_design}
The flexibility of convolution on native points presents a significant computational challenge. To make this high-fidelity design feasible at scale, we introduce a performant systematic design that spans from the fundamental computation abstraction to a highly-optimized GPU kernel.

\subsubsection{Computational Abstraction: MVMR}
 Same as convolution on 2D images, convolution on native points is an instance of~\Eref{eq:generalized_conv}. A tempting strategy for efficient computation of it is to adapt the \emph{im2col} like memory materialization technique once was used from image convolution, to lower the computation into a large GEMM problem. The extra memory usage introduced by materialization of the kernels is $K \times C_{\text{out}} \times C_{\text{in}}$, where $K$ is the number of kernel matrices to use, and $K=t^3$ if a $t$ resolution convolution kernel on each spatial axis is used. While extra memory usage for materializing the kernels are manageable, the materialization of neighborhoods would result in extra memory usage of $N_{\text{in}} \times K \times C_{\text{in}}$, where $N_{\text{in}}$ is the size of the input point cloud, which is $K$ times larger than the original data. Note that, the materialization of neighborhoods would ``fill in" every neighborhood to a uniform maximum size with padding, causing a memory footprint increase far greater than that seen in image convolution and rendering the strategy impractical. Clearly, the strategy that was effective in convolution on 2D images is certainly not acceptable in convolution for native points in terms of extra memory usage, not to mention the latency introduced by the involved memory read and write.

Considering that point cloud is inherently irregular, to the best of our knowledge, there is no special structure to leverage for an alternative strategy in lowering the computation of convolution on native points into a single performant primitive. Inspired by the way cuDNN~\cite{chetlur2014cudnn} addresses the extra memory usage problem and its effectiveness, we hypothesize that a computation abstraction that literally follows the general~\Eref{eq:generalized_conv} formulation has the potential of being performant while achieving zero extra memory usage.

We abstract the computation of~\Eref{eq:generalized_conv} as a Matrix-Vector Multiplication and Reduction (MVMR) problem, with the observation that MVMR naturally decomposes into two distinct components: the $\mathbf{W}_k \times \mathbf{F}^{\text{in}}_j$ component is a Matrix-Vector Multiplication(MVM), which itself is a popular operation with rich studies on achieving performant computation on GPUs, and the rest component is a Reduction --- another example of cases with performant solutions. We develop a unified efficient GPU algorithm by taking inspirations from existing studies on the two sub-problems.
\begin{figure}[!t]
    \centering
    \includegraphics[width=\linewidth]{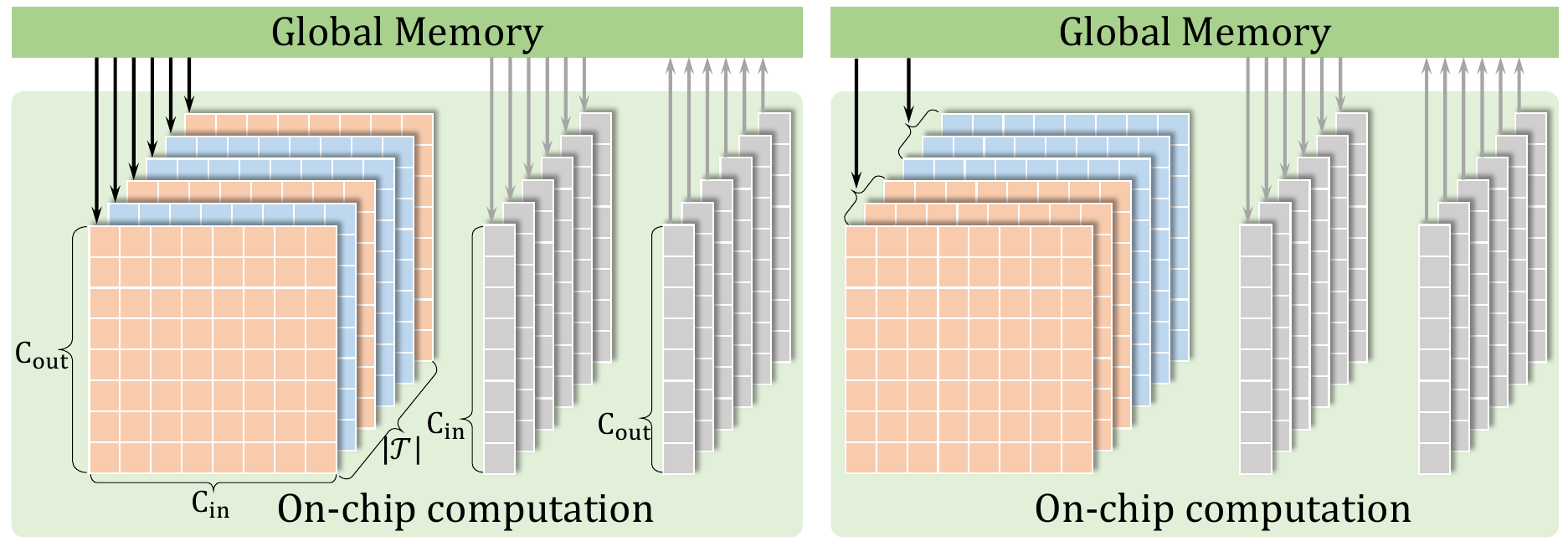}
    \vspace{-5mm}
    \caption{A brute-force MVM computation inefficiently reads $\mathbf{W}_k \in \mathbb{R}^{C_{\text{out}} \times C_{\text{in}}}$ from global memory $\left| \mathcal{T} \right|$ times—once for every triplet(left). Sorting the triplets by $k$ optimizes this. Ideally, each unique $\mathbf{W}_k$ is loaded just once into on-chip memory and reused for all its associated computations(right).}
    \label{fig:memory_access}
    \vspace{-5mm}
\end{figure}

\vspace{-3mm}
\subsubsection{Efficient GPU Algorithm of MVMR}
First of all, a native MVM implementation is required because MVMR involves many small MVM computations. Calling a standard library routine for each computation individually from an outer loop would incur prohibitive launch overhead. Second, while the amount of multiplication and addition is fixed, the key to a performant implementation of MVMR lies in optimizing memory access patterns. GPU performance is critically dependent on:
\begin{inparaenum}[a)]
\item data locality,~\emph{i.e.}, how much data is accessed from fast, on-chip resources versus the much slower, off-chip global memory.
\item coalesced memory access, where threads within a hardware warp access contiguous memory locations to maximize bandwidth.
\end{inparaenum}

The output of the convolution is a feature tensor $\mathbf{F}^{\text{out}} \in \mathbb{R}^{N_{\text{out}} \times C_{\text{out}}}$, where $N_{\text{out}}$ is the size of the output point cloud.
It is computed from three inputs to the convolution operator: 
\begin{inparaenum}[1)]
\item the input feature tensor $\mathbf{F}^{\text{in}} \in \mathbb{R}^{N_{\text{in}} \times C_{\text{in}}}$;
\item the weight matrix $\mathbf{W} \in \mathbb{R}^{K \times C_{\text{out}} \times C_{\text{in}}}$;
\item the collection of triplets $\mathcal{T} = \{(i, j, k)\}$, constructed based on the convolution definition, where  $i$, $j$ and $k$ are integers in range $[0, N_{\text{out}})$, $[0, N_{\text{in}})$, and $[0, K)$, respectively.
\end{inparaenum}
The computational interactions among $\mathbf{F}^{\text{out}}$, $\mathbf{W}$ and $\mathbf{F}^{\text{in}}$ are completely depicted by the triplets $\mathcal{T}$. In most cases, $K \ll N_{\text{in}} \simeq N_{\text{out}} \ll \left| \mathcal{T} \right|$.

Note that the process of each triplet is independent of others, as long as the conflict on Reduction write to $\mathbf{F}^{\text{out}}$ is governed by atomic primitive. Therefore, there is a straightforward parallelism strategy to accomplish the computation of all the triplets: let each thread handle the computation of one triplet. While this strategy seems ideal in terms of parallelism, significant amount of slow global memory access is involved, as illustrated in~\Fref{fig:memory_access} (left):
\vspace{-2mm}
\begin{equation}
\label{eq:memory_brute_force}
\left| \mathcal{T} \right| \times (\underbrace{C_{\text{out}} \times C_{\text{in}}}_{\text{Read from }\mathbf{W}} + \underbrace{C_{\text{in}}}_{\text{Read from } \mathbf{F}^{\text{in}}} +  \underbrace{C_{\text{out}}}_{\text{Atomic Write to } \mathbf{F}^{\text{out}}}).
\end{equation}
Clearly, the global memory read from $\mathbf{W}$ is a bottleneck, and therefore, this access pattern must be addressed to improve performance. Mathematically, the final results are invariant to the processing order of the triplets\footnote{Practically, the results may vary with order due to numerical effects (\emph{e.g.}, floating-point arithmetic).}, but a sorting of the triplets could significantly change the memory access patterns, as illustrated in~\Fref{fig:memory_access} (right). More specifically, given the triplet list $\mathcal{T} = \{(i, j, k)\}$ sorted by $k$, if it is split into consecutive groups of length $L$, based on Pigeonhole Principle, \emph{most} of the groups contain triplets that share the same $k$ value.
More specifically, in the case of $K \ll \left| \mathcal{T} \right|$, a practical approximations of the expectation number of unique $k$ in each group is $1+\frac{L \times K}{\left| \mathcal{T} \right|}$\footnote{This is a classic problem that combines order statistics with a variant of the coupon collector's problem. The expected number of unique values in a single group is $1+(L-1) \left[ \frac{K(1-{(1-\frac{1}{K})}^{\left| \mathcal{T} \right|})-1}{\left| \mathcal{T} \right|-1} \right]$.}, which approaches 1, when $L \times K \ll \left| \mathcal{T} \right|$.
Due to this reason, when each of such group is processed together, \emph{almost} only one global memory read of $\mathbf{W}$ is required for each group, and the overall global memory access could be effectively reduce to $\mathcal{O}(\frac{1}{L})$ of that in~\Eref{eq:memory_brute_force}:
\vspace{-3mm}
\begin{equation}
\label{eq:memory_grouped}
\frac{\left| \mathcal{T} \right|}{L} \times (\underbrace{C_{\text{out}} \times C_{\text{in}}}_{\text{Read from }\mathbf{W}} + \underbrace{L \times C_{\text{in}}}_{\text{Read from } \mathbf{F}^{\text{in}}} + \underbrace{L\times C_{\text{out}}}_{\text{Atomic Write to } \mathbf{F}^{\text{out}}}).
\end{equation}
Beyond the option of sorting $\mathcal{T}$ by $k$, there are two alternatives: sorting by $i$ and sorting by $j$. Following the analysis of the saving introduced from sorting by $k$, they would result in saving read from $\mathbf{F}^{\text{in}}$ and atomic write to $\mathbf{F}^{\text{out}}$, respectively. However, both of these savings are less effective, as:
\begin{inparaenum}[1)]
\item the amount to load from $\mathbf{F}^{\text{in}}$ or atomic write to $\mathbf{F}^{\text{out}}$ is often orders of magnitude smaller than the read from $\mathbf{W}$;
\item the expectation number of unique $i$ and $j$ in each group is $1+\frac{L \times N_{\text{out}}}{\left| \mathcal{T} \right|}$ and $1+\frac{L \times N_{\text{in}}}{\left| \mathcal{T} \right|}$, respectively, --- not as ideal as the $1+\frac{L \times K}{\left| \mathcal{T} \right|}$ in the sorting by $k$ case, as $K \ll N_{\text{in}} \simeq N_{\text{out}}$.
\end{inparaenum}
Our analysis coincides with our profiling, thus we choose to sort by $k$. Nevertheless, the above analysis is based on typical configurations, in terms of unknown configurations, an auto tuning mechanism could be used to select the index to be sorted by, before the massive executions.
\begin{algorithm}[!t]
    \begin{spacing}{0.85}
    \small
    \caption{MVMR Kernel for Computing~\Eref{eq:generalized_conv}}
    \label{alg:mvmr_kernel}
    \begin{algorithmic}[1]
        \renewcommand{\algorithmicrequire}{\textbf{Inputs:}}
        \Require $\mathbf{W}$, $\mathbf{F}^{\text{in}}$, $\mathcal{T}^{L}$.
        
        \renewcommand{\algorithmicensure}{\textbf{Output:}}
        \Ensure $\mathbf{F}^{\text{out}}$.

        \algnewcommand{\AtomicAdd}[2]{\textbf{atomicAdd}(#1, #2)}
        
        \State{$(\tilde{i}, \tilde{j}, \tilde{k}) \leftarrow \mathcal{T}^{L}_0$} \Comment{initialize with the first triplet.}
        \State $\red{\tilde{W}} \leftarrow \mathbf{W}_{\tilde{k}}$, $\red{\tilde{F}^{\text{in}}} \leftarrow \mathbf{F}^{\text{in}}_{\tilde{j}}$ \Comment{read from global memory.}
        \State $\red{\tilde{F}^{\text{out}}} \leftarrow \red{\tilde{W}} \times \red{\tilde{F}^{\text{in}}}$ \Comment{fast on-chip computation.}
        \ForAll{$(i, j, k)$ in $\mathcal{T}^{L}_{[1,2, ...,  L-1]}$}
            \If{$j \neq \tilde{j}$}
                \State $\tilde{j} \leftarrow j$, $\red{\tilde{F}^{\text{in}}} \leftarrow \mathbf{F}^{\text{in}}_{\tilde{j}}$ \Comment{read only if necessary.}
            \EndIf{}
            
            \If{$k \neq \tilde{k}$}
                \State $\tilde{k} \leftarrow k$, $\red{\tilde{W}} \leftarrow \mathbf{W}_{\tilde{k}}$ \Comment{read only if necessary.}
            \EndIf{}

            \If{$i \neq \tilde{i}$}
                \State \AtomicAdd{$\red{\tilde{F}^{\text{out}}}$}{$\mathbf{F}^{\text{out}}_{\tilde{i}}$} \Comment{only if necessary.}
                \State $\tilde{i} \leftarrow i$, $\red{\tilde{F}^{\text{out}}} \leftarrow \red{\tilde{W}} \times \red{\tilde{F}^{\text{in}}}$ \Comment{on-chip, fast.}
            \Else
                \State $\red{\tilde{F}^{\text{out}}} \leftarrow \red{\tilde{F}^{\text{out}}} + \red{\tilde{W}} \times \red{\tilde{F}^{\text{in}}}$ \Comment{on-chip, fast.}
            \EndIf{}
        \EndFor
        \State \AtomicAdd{$\red{\tilde{F}^{\text{out}}}$}{$\mathbf{F}^{\text{out}}_{\tilde{i}}$} 
    \end{algorithmic}
    \end{spacing}
\end{algorithm}

While the strategy of grouped processing on the triplet list sorted by $k$ is effective in addressing the critical data locality issue, a naive parallelism strategy by letting each thread handle the computation of one group is impractical as there is often not enough on-chip memory to afford an entire $C_{\text{out}} \times C_{\text{in}}$ matrix for each thread. We discern that the MVM sub-problem in MVMR is an extreme case of an already specialized Tall-and-Skinny Matrix-Matrix Multiplication problem, by viewing a vector as a one column matrix, for which an efficient GPU algorithm is proposed in~\cite{RIVERA202170}. We refer the readers to~\cite{RIVERA202170} for the design considerations and details that could be easily adopted into our MVM sub-problem. Heavily based on this, a practical and coalesced memory access friendly parallelism strategy that integrates the aforementioned grouping logic is proposed. This strategy divides each of the $C_{\text{out}} \times C_{\text{in}}$ kernel matrix into $B_{\text{out}} \times B_{\text{in}}$ blocks, respectively, and assigns the computation related to each block of a group's MVM computation to a warp of threads, rather than assigning the entire computation to a single thread.

We summarize our algorithm for efficient computation of MVMR in~\Aref{alg:mvmr_kernel}, where $\mathcal{T}^{L} = \{(i, j, k)\}$ are the $L$ triplets assigned to a warp of threads, and on-chip resources are denoted as $\red{\tilde{W}} \in \mathbb{R}^{B_{\text{out}} \times B_{\text{in}}}$, $\red{\tilde{F}^{\text{in}}} \in \mathbb{R}^{B_{\text{in}}}$, and $\red{\tilde{F}^{\text{out}}} \in \mathbb{R}^{B_{\text{out}}}$. Note that effective global memory access saving are supported for all the three sorting options in~\Aref{alg:mvmr_kernel}. An implementation based on Triton~\cite{tillet2019triton} is provided in supplementary materials. There are three hyperparameters in our algorithm, $L$, $B_{\text{out}}$ and $B_{\text{in}}$, we use 128, 32 and 32, respectively, in all of our experiments.

Note that there is zero extra memory usage, as all the computation are native on the input tensors, without resorting to any intermediate global memory --- an achieved goal similar to that in cuDNN~\cite{chetlur2014cudnn} on the implementation of efficient convolution for dense and regular data. The one and only ``extra'' computation in our algorithm is the sorting of the triplets by $k$, which is an highly optimized parallel algorithm on GPU, the latency of which is negligible.

\subsubsection{Efficient Gradient Computation}
An efficient backward pass is critical for performant model training and finetuning. By applying the chain rule to our generalized convolution as described in~\Eref{eq:generalized_conv}, two required gradients in distinct computational patterns are revealed, both of which demand efficient computation.

First, the gradient with respect to the input feature $\mathbf{F}^{\text{in}}_j$ is an accumulation of transformed output gradients:
\vspace{-2mm}
\begin{equation}
    \nabla_{\mathbf{F}^{\text{in}}_j} \mathcal{L} = \sum_{(i, j, k) \in \mathcal{T}} {\mathbf{W}_k}^T \times \nabla_{\mathbf{F}^{\text{out}}_i} \mathcal{L}.
\end{equation}
This operation mirrors the structure of the forward pass, where a set of matrices (now the transposed weights, ${\mathbf{W}_k}^T$) are multiplied by a set of vectors (the incoming gradients, $\nabla_{\mathbf{F}^{\text{out}}_i} \mathcal{L}$) and reduced. Consequently, this computation can be framed as an MVMR problem, allowing us to leverage the very same highly-optimized kernel designed for the forward pass, ensuring high efficiency.

Second, computing the gradient with respect to a weight kernel $\mathbf{W}_k$ involves summing the outer products of the upstream output gradients and the corresponding input feature vectors over all associated triplets:
\vspace{-2mm}
\begin{equation}
\label{eq:bw_kernel}
    \nabla_{\mathbf{W}_k}\mathcal{L} = \sum_{(i, j, k) \in \mathcal{T}} \nabla_{\mathbf{F}^{\text{out}}_i}\mathcal{L} \otimes \mathbf{F}^{\text{in}}_j.
\end{equation}
This structure follows a different pattern, which we abstract as Vector-Vector Outer product and Reduction (VVOR). For this, we develop a second dedicated GPU kernel, following similar principles used to develop the kernel for MVMR, which efficiently computes these outer products and reduces them into the final weight gradients. The algorithm details, as well as its Triton based implementation, of VVOR, are provided in the supplementary materials.

By implementing the backward pass with these two specialized, performant kernels—reusing MVMR for feature gradients and introducing VVOR for weight gradients—we ensure that the entire training process is computationally efficient, completing our holistic system design.
\vspace{-2mm}

%% file: sec/4_exps.tex
\section{Experiments}
Our experiments are designed to validate that PointCNN++ delivers both high performance and high fidelity. We first detail our implementation in~\Sref{sec:implementation}, then use micro-benchmarks to quantify its GPU memory usage and computational efficiency with extensive comparisons against the performance prioritized voxel-based methods in~\Sref{sec:ablation}. Finally, we demonstrate its superior accuracy and generalization through two downstream tasks: the geometrically sensitive task of point cloud registration in~\Sref{sec:registration} and the large-scale scene understanding task of semantic segmentation in~\Sref{sec:segmentation}.
\vspace{-1mm}
\subsection{Implementation Details}
\label{sec:implementation}
We represent point clouds as jagged tensors same as that in~\cite{williams2024fvdb}, and also highlight specific architectural refinements. While not the core novelty of our work, these choices diverge from common practices and offer notable benefits for quality and robustness, which we hope will prove valuable to the community. We opt for a fixed radius search over a fixed-number (K-Nearest Neighbors, or KNN) search, as its spatially-local receptive field is better suited for convolutional learning, whereas KNN is often a choice imposed by architectural limitations. For sampling, we employ a voxel-based downsampling that, by not snapping points to voxel centers, better preserves thin structures and sparsely captured regions than random downsampling with negligible latency (contradicting the claim in~\cite{hu2020randla} that random sampling is essential for efficiency), and we use the original pre-downsampled points for efficient upsampling. While these operations can be lowered into spatial lookups as in Open3D~\cite{zhou2018open3d} with the underlying ASH~\cite{dong2022ash} engine based on hash map, our empirical findings led us to implement a more robust and competitively fast mechanism built upon two highly-optimized GPU primitives: sorting and searching. The implementation of these operations will be open sourced alongside our core convolution operator.

\begin{figure}[t]
    \centering
    \includegraphics[width=0.98\linewidth]{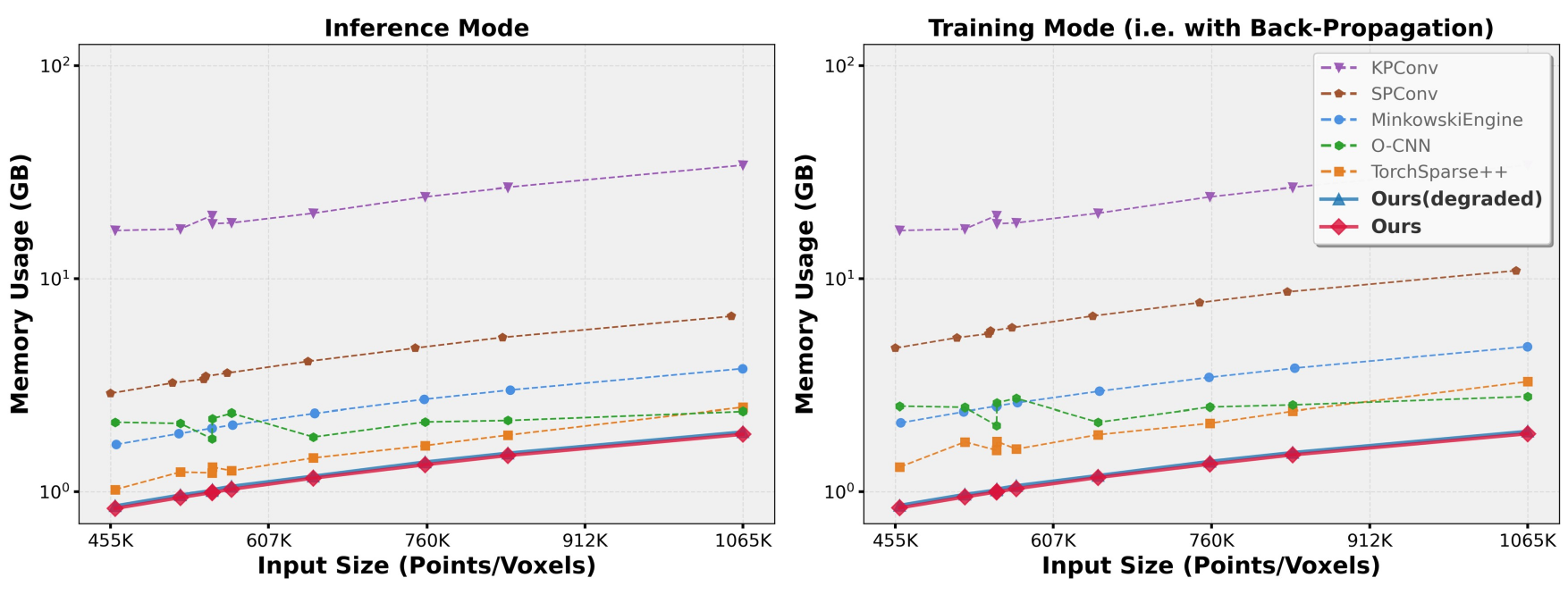}
    \vspace{-4mm}
    \caption{Memory usage comparison of one convolution layer.}
    \label{fig:convm}
    \vspace{-4mm}
\end{figure}
\begin{figure}[t]
    \centering
    \includegraphics[width=0.98\linewidth]{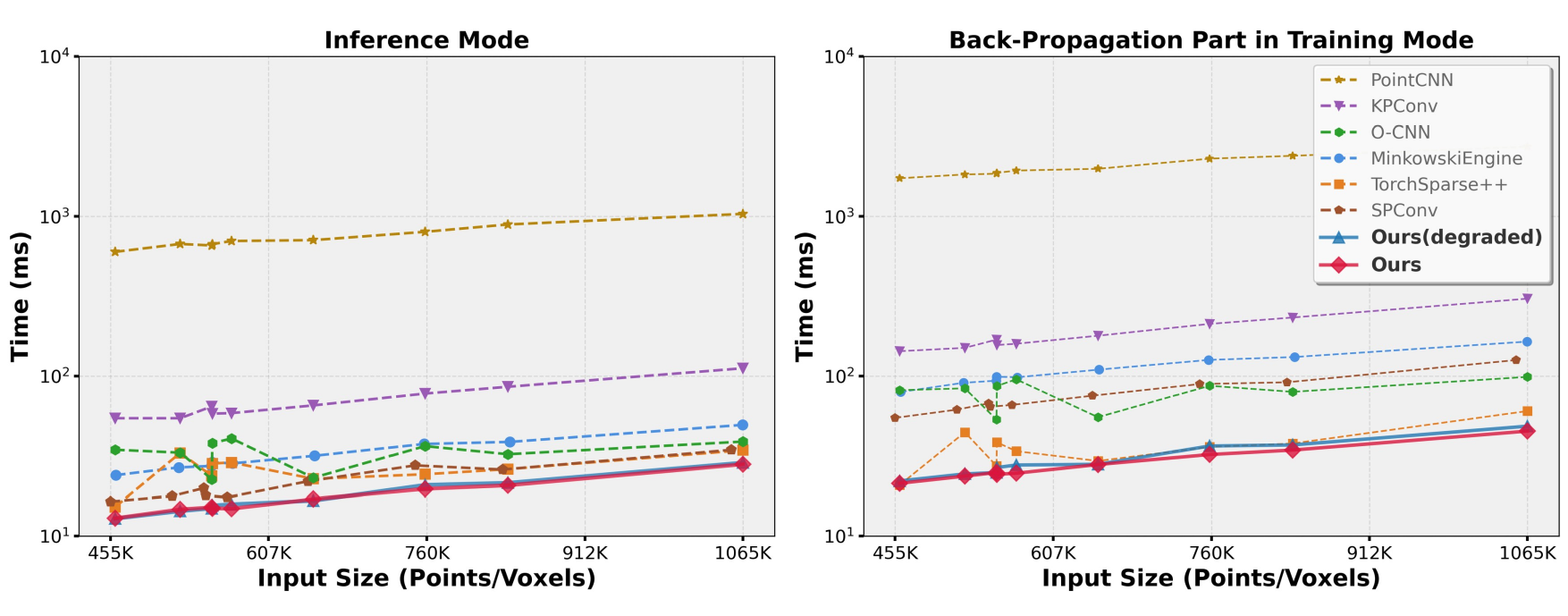}
    \vspace{-4mm}
    \caption{Performance comparison of one convolution layer.}
    \label{fig:convt}
    \vspace{-5mm}
\end{figure}
\vspace{-1mm}
\subsection{Performance Study}
\label{sec:ablation}
\vspace{-1mm}
This section is dedicated to analyzing the computational performance of our method. We establish its efficiency and scalability by benchmarking its memory footprint and latency against other representative methods on the isolated convolution operator level, as well as the timing for the end-to-end forward and backward passes\footnote{See supplementary material for performance benchmarks on various GPUs and scalability limits.}.
\vspace{1mm}
\begin{figure}[t]
    \centering
    \includegraphics[width=\linewidth]{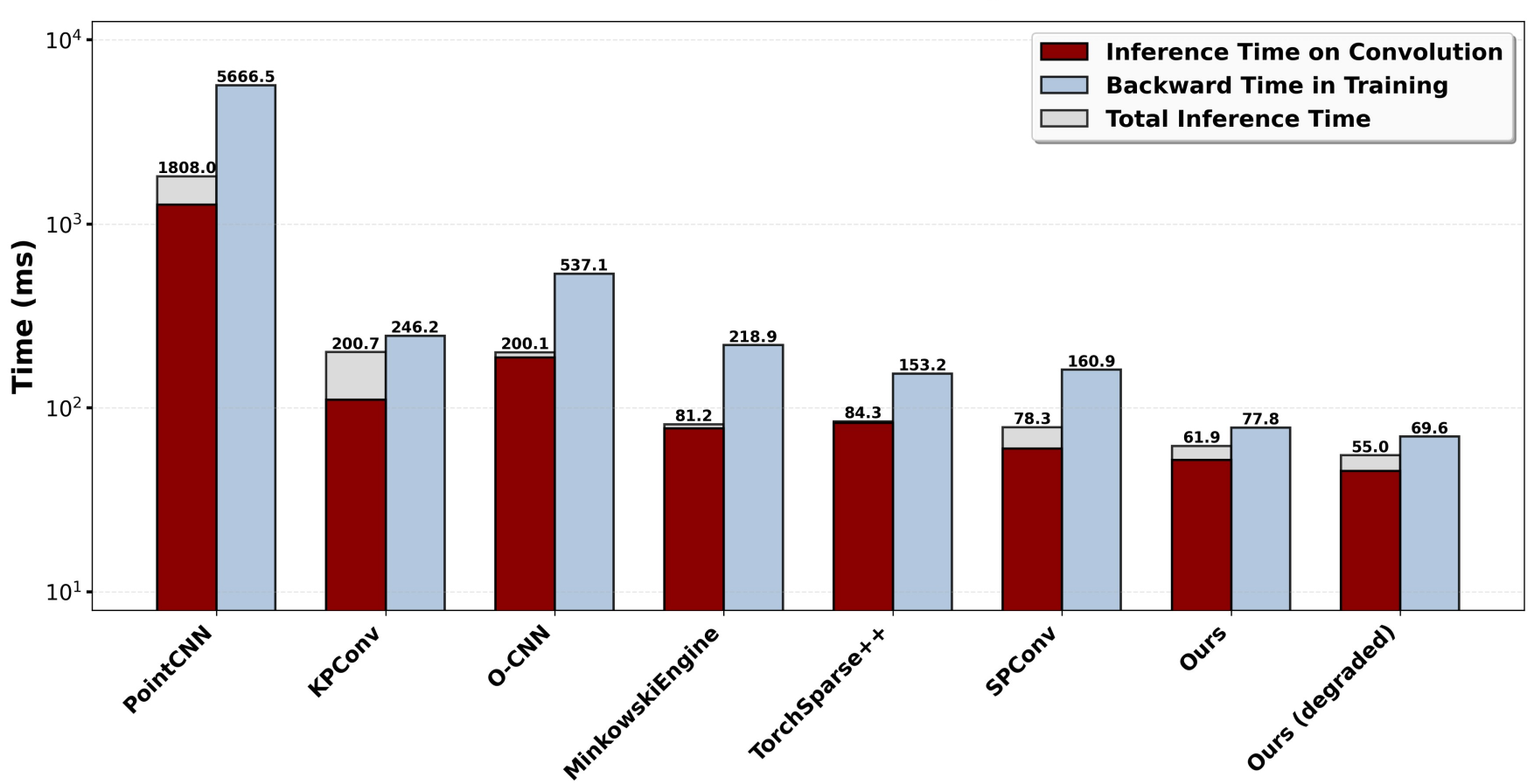}
    \vspace{-7mm}
    \caption{Performance of various convolutional backbones for 3D learning assembled in an identical ResNet-18~\cite{he2016deep} architecture.}
    \label{fig:fwbwt}
    \vspace{-3mm}
\end{figure}
\begin{table}[t]
\centering
\caption{Quantitative comparison for point cloud registration on the KITTI.
The best and second-best results are in \textcolor{red}{red} and \textcolor{blue}{blue}. Architecture abbreviations: MkEngine (MinkowskiEngine), KP+Attn (KPConv with Attention), KP+GCN (KPConv with Graph Convolutional Network), and PNpp+Attn (PointNet++ with Attention), with task-specific designs of each method list in \textit{italic}.}
\label{tab:registration_kitti}
\resizebox{0.45\textwidth}{!}{
\begin{tabular}{l c cc cc cc c}
\toprule
\multirow{2}{*}{\textbf{Method}} & \multirow{2}{*}{\textbf{Arch.}} & \multicolumn{2}{c}{\textbf{RTE(m)}~$\downarrow$} & \multicolumn{2}{c}{\textbf{RRE($^\circ$)}~$\downarrow$} & \multicolumn{1}{c}{\textbf{Recall(\%)~$\uparrow$}} & \multirow{2}{*}{\textbf{\textbf{Param}}}\\
\cmidrule(lr){3-4} \cmidrule(lr){5-6} \cmidrule(lr){7-7}
& & \textbf{Mean} & \textbf{Std} & \textbf{Mean} & \textbf{Std} &  \textbf{@0.2m,1$^\circ$} \\
\midrule
\multirow{2}{*}{\begin{tabular}[c]{@{}l@{}}FCGF\\{\scriptsize (2019)}\end{tabular}} & \textbf{MkEngine} & \multirow{2}{*}{0.36} & \multirow{2}{*}{0.133} & \multirow{2}{*}{0.110} & \multirow{2}{*}{0.41} & \multirow{2}{*}{85.2} & \multirow{2}{*}{8.75M} \\
& \textcolor{gray}{\scriptsize \textit{ResUNet}} & & & & & & \\
\cmidrule{2-2}
\multirow{2}{*}{\begin{tabular}[c]{@{}l@{}}DGR\\{\scriptsize (2020)}\end{tabular}} & \textbf{MkEngine} & \multirow{2}{*}{0.35} & \multirow{2}{*}{0.072} & \multirow{2}{*}{0.090} & \multirow{2}{*}{0.36} & \multirow{2}{*}{92.1} & \multirow{2}{*}{244.68M} \\
& \textcolor{gray}{\scriptsize \textit{IR prediction}} & & & & & & \\
\cmidrule{2-2}
\multirow{2}{*}{\begin{tabular}[c]{@{}l@{}}CoFiNet\\{\scriptsize (2021)}\end{tabular}} & \textbf{KP+Attn} & \multirow{2}{*}{0.39} & \multirow{2}{*}{0.065} & \multirow{2}{*}{0.091} & \multirow{2}{*}{0.39} & \multirow{2}{*}{89.4} & \multirow{2}{*}{5.48M} \\
& \textcolor{gray}{\scriptsize \textit{Coarse2Fine}} & & & & & & \\
\cmidrule{2-2}
\multirow{2}{*}{\begin{tabular}[c]{@{}l@{}}Predator\\{\scriptsize (2021)}\end{tabular}} & \textbf{KP+GCN} & \multirow{2}{*}{0.35} & \multirow{2}{*}{0.063} & \multirow{2}{*}{0.081} & \multirow{2}{*}{0.27} & \multirow{2}{*}{93.9} & \multirow{2}{*}{158.42M} \\
& \textcolor{gray}{\scriptsize \textit{Overlap pred.}} & & & & & & \\
\cmidrule{2-2}
\multirow{2}{*}{\begin{tabular}[c]{@{}l@{}}GeoTrans\\{\scriptsize (2022)}\end{tabular}} & \textbf{KP+Attn} & \multirow{2}{*}{0.38} & \multirow{2}{*}{0.068} & \multirow{2}{*}{0.096} & \multirow{2}{*}{0.29} & \multirow{2}{*}{85.6} & \multirow{2}{*}{25.50M} \\
& \textcolor{gray}{\scriptsize \textit{Geom emb.}} & & & & & & \\
\cmidrule{2-2}
\multirow{2}{*}{\begin{tabular}[c]{@{}l@{}}Regformer\\{\scriptsize (2023)}\end{tabular}} & \textbf{PNpp+Attn} & \multirow{2}{*}{\textcolor{blue}{\textbf{0.22}}} & \multirow{2}{*}{0.077} & \multirow{2}{*}{\textcolor{red}{\textbf{0.058}}} & \multirow{2}{*}{0.28} & \multirow{2}{*}{\textcolor{blue}{\textbf{94.6}}} & \multirow{2}{*}{3.12M} \\
& \textcolor{gray}{\scriptsize \textit{Proj-aware}} & & & & & & \\
\cmidrule{2-2}
\multirow{2}{*}{\begin{tabular}[c]{@{}l@{}}UMEReg\\{\scriptsize (2024)}\end{tabular}} & \textbf{MkEngine} & \multirow{2}{*}{0.49} & \multirow{2}{*}{\textcolor{red}{\textbf{0.023}}} & \multirow{2}{*}{0.490} & \multirow{2}{*}{\textcolor{blue}{\textbf{0.21}}} & \multirow{2}{*}{79.6} & \multirow{2}{*}{7.17M} \\
& \textcolor{gray}{\scriptsize \textit{UME + SEM}} & & & & & & \\
\midrule
\multirow{2}{*}{\begin{tabular}[c]{@{}l@{}}\textbf{Ours}\\{\scriptsize \textbf{(2025)}}\end{tabular}} & \textbf{PointCNN++} & \multirow{2}{*}{\textcolor{red}{\textbf{0.19}}} & \multirow{2}{*}{\textcolor{blue}{\textbf{0.03}}} & \multirow{2}{*}{\textcolor{blue}{\textbf{0.060}}} & \multirow{2}{*}{\textcolor{red}{\textbf{0.10}}} & \multirow{2}{*}{\textcolor{red}{\textbf{99.8}}} & \multirow{2}{*}{8.75M} \\
& \textcolor{gray}{\scriptsize \textit{ResUNet}} & & & & & & \\
\bottomrule
\end{tabular}
}
\vspace{-5mm}
\end{table}
\vspace{-1mm}
\begin{table*}[t]
\centering
\caption{Quantitative comparison on 3DMatch across different input point densities. We report Registration Recall (RR~$\uparrow$), Feature Matching Recall (FMR~$\uparrow$), and Inlier Ratio (IR~$\uparrow$), all in percentages (\%). The best results are in red bold, and the second best in blue bold.}
\vspace{-3mm}
\label{tab:registration_3dmatch}
\resizebox{0.8\textwidth}{!}{
\begin{tabular}{l ccc| ccc |ccc| ccc| cccc}
\toprule
\multirow{2}{*}{\textbf{Method}} & \multicolumn{3}{c}{\textbf{5000 Pts}} & \multicolumn{3}{c}{\textbf{2500 Pts}} & \multicolumn{3}{c}{\textbf{1000 Pts}} & \multicolumn{3}{c}{\textbf{500 Pts}} & \multicolumn{3}{c}{\textbf{250 Pts}}  \\
\cmidrule(lr){2-4} \cmidrule(lr){5-7} \cmidrule(lr){8-10} \cmidrule(lr){11-13} \cmidrule(lr){14-16}
& \textbf{RR} & \textbf{FMR} & \textbf{IR} & \textbf{RR} & \textbf{FMR} & \textbf{IR} & \textbf{RR} & \textbf{FMR} & \textbf{IR} & \textbf{RR} & \textbf{FMR} & \textbf{IR} & \textbf{RR} & \textbf{FMR} & \textbf{IR} & \\
\midrule
FCGF(2019)               & 85.1 & 97.4 & 52.8 & 84.7 & 97.3 & 51.1 & 83.3 & 97.0 & 46.7 & 81.6 & 96.7 & 41.5 & 71.4 & 96.6 & 34.1  \\
CoFiNet(2021)            & 89.3 & 98.1 & 49.8 & 88.9 & {\color{blue}\textbf{98.3}} & 51.2 & 88.4 & {\color{blue}\textbf{98.1}} & 51.9 & 87.4 & {\color{blue}\textbf{98.2}} & 52.2 & 87.0 & {\color{blue}\textbf{98.3}} & 52.2 \\
Predator(2021)           & 89.0 & 96.6 & 58.0 & 89.9 & 96.6 & 58.4 & 90.6 & 96.5 & 57.1 & 88.5 & 96.3 & 54.1 & 86.6 & 96.5 & 49.3 \\
GeoTrans(2023)     & {\color{red}\textbf{91.4}} & {\color{blue}\textbf{97.9}} & {\color{red}\textbf{70.5}} & {\color{red}\textbf{91.1}} & 98.0 & {\color{red}\textbf{72.9}} & {\color{red}\textbf{92.0}} & {\color{blue}\textbf{98.1}} & {\color{red}\textbf{75.2}} & {\color{red}\textbf{91.7}} & {\color{blue}\textbf{98.2}} & {\color{red}\textbf{79.8}} & {\color{red}\textbf{91.2}} & {\color{blue}\textbf{98.3}} & {\color{red}\textbf{84.6}} \\
\midrule
\textbf{Ours}            & {\color{blue}\textbf{90.3}} & {\color{red}\textbf{99.3}} & {\color{blue}\textbf{58.2}} & {\color{blue}\textbf{90.2}} & {\color{red}\textbf{99.3}} & {\color{blue}\textbf{61.4}} & {\color{blue}\textbf{89.2}} & {\color{red}\textbf{99.3}} & {\color{blue}\textbf{62.5}} & {\color{blue}\textbf{89.1}} & {\color{red}\textbf{99.3}} & {\color{blue}\textbf{63.4}} & {\color{blue}\textbf{88.3}} & {\color{red}\textbf{99.1}} & {\color{blue}\textbf{64.1}}  \\
\bottomrule
\end{tabular}
}
\end{table*}\\
\noindent\textbf{Experimental Setup.} To ensure a fair and controlled comparison, all backbones used in this study are built upon a ResNet-18~\cite{he2016deep} architecture. 
For the operator-level benchmarks, we use a standard convolution configuration ($C_\text{in}=64,C_\text{out}=128,K=3^3$).
We benchmark our method against several representative point-based~\cite{li2018pointcnn,thomas2019kpconv} and voxel-based~\cite{wang2017cnn,choy20194d,spconv2022,tangandyang2023torchsparse} backbones.
Crucially, we also include a variant named `Ours (degraded)', as described in~\Sref{sec:point_centric_conv}. This allows us to directly quantify the performance difference between our approach and a traditional voxel-based paradigm within the same framework.
For both latency and memory analysis, we use 10 large-scale scenes from the S3DIS dataset~\cite{Armeni_2016_CVPR}. Timings and peak GPU memory consumption are recorded on a single NVIDIA RTX 4090 GPU. To ensure stability, all reported results are averaged over 10 independent runs per scene.
%

\noindent\textbf{Memory Analysis.} The memory usage results in~\Fref{fig:convm} provide compelling validation for the ``nativeness'' of our method. Our operator demonstrates unparalleled efficiency, consistently consuming the least GPU memory by a significant margin than existing voxel-based methods and over an order of magnitude less than existing point-based approaches\footnote{The implementation of PointCNN~\cite{li2018pointcnn} is based on TensorFlow, while all the rest are based on PyTorch. We exclude the memory usage comparison with it, as the differences on the memory allocation strategy of these two frameworks prevent a fair direct comparison.}. This advantage is a direct consequence of our kernel's design, which operates with a zero extra memory footprint. It fundamentally bypasses the need for the memory-intensive padding or tensor materialization that inflates the memory consumption of competing approaches. These results serve as a powerful testament to the efficacy and superior architectural design of our native operator.

\noindent\textbf{Latency Analysis.} 
Our analysis begins with the core convolution operator (\Fref{fig:convt}), where our method substantially outperforms leading point-based and voxel-based alternatives. As shown in~\Fref{fig:fwbwt}, with a high-quality implementation of other utilities, as described in~\Sref{sec:implementation}, the convolution operations dominate network latency in our method, thus the operator-level advantage translates directly to powerful end-to-end performance, which is faster than all baselines in both forward and backward passes. Surprisingly, our method, when degraded into a voxel-based variant as described in~\Sref{sec:point_centric_conv}, demonstrated an additional speed improvement over the full method.\footnote{We did not expect this, as we anticipated statistically equivalent performance. While the definitive cause for this speed-up remains unknown, our analysis showed that the variant's kernel matrix usage is more concentrated on spatially 'central' kernels. We conjecture that this access pattern could lead to improved cache performance, though this requires further investigation.}
\vspace{-1mm}
\subsection{Downstream Task: Point Cloud Registration}
\label{sec:registration}
Our primary goal here is to investigate a simple yet critical question: \red{\textbf{\textit{can solely replacing a standard voxel-based backbone with our geometrically faithful PointCNN++ counterpart yield significant performance gains?}}}

To this end, we deliberately avoid any task-specific tuning.
%
We take the classic FCGF~\cite{choy2019fully} architecture and simply replace its MinkowskiEngine-based backbone with our PointCNN++ implementation. All other components
remain identical to the original FCGF.
We then compare this straightforwardly upgraded model against a series of recent state-of-the-art methods, including DGR~\cite{choy2020deep}, Predator~\cite{huang2021predator}, CoFiNet~\cite{yu2021cofinet}, GeoTrans~\cite{qin2023geotransformer}, Regformer~\cite{liu2023regformer}, and UMEReg~\cite{haitman2024umeregrobust}. These methods often achieve high performance through highly-specific mechanisms and sophisticated designs.

\vspace{-1mm}
\begin{figure*}[t]
\centering
\includegraphics[width=\linewidth]{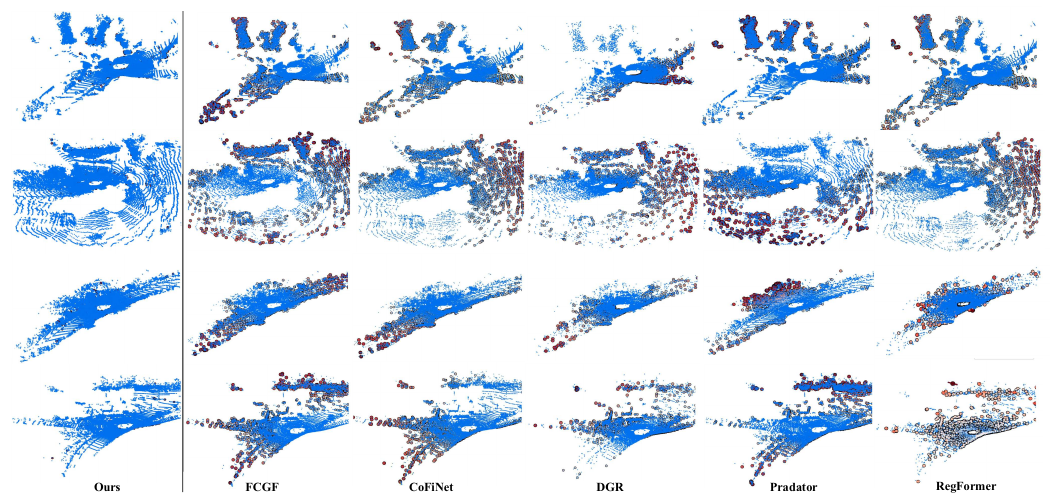}
\caption{Point-wise registration error visualization on KITTI dataset comparing our method with state-of-the-art baselines. The visualization shows errors between ground truth point clouds and transformed source point clouds, where darker red indicates larger errors. (Registration errors below 0.1m are ignored)}
\vspace{-3mm}
\label{fig:kitti_registration_baselines}
\end{figure*}

%
%
In contrast, our approach brings only a better engine to a vintage chassis, yet, as we will show, proves to be remarkably competitive. Exploring the fusion of our backbone with advanced modules
is left for future work.

We conduct evaluations on two standard benchmarks that represent distinct environments.
KITTI Odometry~\cite{geiger2012we} is a widely-used outdoor dataset from an autonomous driving platform, featuring sparse LiDAR scans. 
%
3DMatch~\cite{zeng20173dmatch} is a large-scale indoor dataset composed of RGB-D scans from various scenes. 
%
Following UMEReg~\cite{haitman2024umeregrobust}, we report Relative Translation Error (RTE), Relative Rotation Error (RRE) and Registration Recall (RR) on the KITTI.
On the 3DMatch, we follow the protocol of CoFiNet~\cite{yu2021cofinet} and use Registration Recall (RR), Feature Matching Recall (FMR), and Inlier Ratio (IR) for evaluation.
%
%

\noindent\textbf{Results on KITTI.} As shown in~\Tref{tab:registration_kitti}, the results on the KITTI dataset powerfully validate the effectiveness of our method in the zero-task specific tuning ``plug-and-play'' setting. By simply replacing the backbone in the classic FCGF architecture, we achieve state-of-the-art performance, nearly halving its RTE to 0.19m and boosting its RR to a near-perfect 99.8\%. Beyond raw accuracy, our model demonstrates unparalleled registration stability, achieving the lowest standard deviations in both translation and rotation by a wide margin. This combination of top-tier accuracy and exceptional consistency proves that our high-fidelity operator produces superior features, leading to quantifiably better and more reliable registrations. The visualization in~\Fref{fig:kitti_registration_baselines} further illustrates the qualitative superiority of our method across different outdoor scenarios.


\noindent\textbf{Results on 3DMatch.} 
As shown in~\Tref{tab:registration_3dmatch}, our method demonstrates compelling results on the 3DMatch benchmark. This is particularly noteworthy as these results are achieved using an old backbone, which is considerably simpler than the bespoke architectures of competitors like GeoTrans~\cite{qin2023geotransformer}. It indicates that our operator is powerful enough on its own to elevate a conventional architecture to produce features of state-of-the-art quality. 

\subsection{Downstream Task: Semantic Segmentation}
\label{sec:segmentation}
To evaluate the generalization and versatility of PointCNN++, we further benchmark it on semantic segmentation, a task that requires capturing both fine-grained local geometry and large-scale contextual information.

\noindent\textbf{Experimental Setup.} We conduct a head-to-head comparison on the challenging nuScenes~\cite{caesar2020nuscenesmultimodaldatasetautonomous} semantic segmentation benchmark. Following the methodology in Section~\ref{sec:registration}, we adopt a ``drop-in'' replacement strategy to ensure fairness. We use a ResUNet2 backbone (with MSC pre-training and a batch size of 3) as the reference architecture. The core convolution operators of the backbone are replaced while keeping the rest of the architecture, including the loss functions and training pipeline, identical.

\noindent\textbf{Baselines.} We compare our operator against representative point-based and voxel-based paradigms. Specifically, we include KPConv~\cite{thomas2019kpconv} as the point-based baseline and MinkowskiEngine~\cite{choy20194d} as the voxel-based baseline. This setup allows us to directly measure the efficiency and accuracy gains provided by the PointCNN++ operator within a consistent network framework.

\noindent\textbf{Results and Analysis.} The results are summarized in Table~\ref{tab:nus}. Under the same experimental setting, training KPConv would require approximately 116 days, rendering it impractical for real-world deployment. We therefore only report its memory and per-iteration time measured during the training phase. PointCNN++ achieves the highest performance with an mIoU of 78.2\% and an mAcc of 85.3\%, outperforming the voxel-based MinkowskiEngine by 3.8\% in mIoU . Crucially, PointCNN++ delivers these improvements with a smaller memory footprint (2.43 GB vs. 2.61 GB) and faster execution (0.102s vs. 0.131s per iteration) than the voxel-based baseline. Compared to the point-based KPConv, PointCNN++ reduces memory usage by nearly an order of magnitude and is over 200$\times$ faster per iteration. These results underscore the dual advantage of our method: it preserves the geometric fidelity typically associated with point-based methods while surpassing the computational efficiency of optimized voxel-based frameworks.
\begin{table}[t]
\centering
\caption{Head-to-head comparison on nuScenes semantic segmentation using a ResUNet2 backbone.}
\label{tab:nus}
\resizebox{0.98\linewidth}{!}{
\begin{tabular}{l|cc|cc}
\hline
\textbf{Operator} & \textbf{mIoU (\%)} $\uparrow$ & \textbf{mAcc (\%)} $\uparrow$ & \textbf{Mem (GB)} $\downarrow$ & \textbf{Time (per iter.)} $\downarrow$ \\ \hline
KpConv (point-based) & - & - & 17.03 & 23.82 \\
MinkowskiEngine (voxel-based) & 74.4 & 81.7 & 2.61 & 0.131 \\
\textbf{PointCNN++ (Ours)} & \textbf{78.2} & \textbf{85.3} & \textbf{2.43} & \textbf{0.102} \\ \hline
\end{tabular}
}
\end{table}
\vspace{-3mm}

%% file: sec/5_conclusion.tex
\section{Conclusion}
\vspace{-1mm}
We present PointCNN++, a solution to the persistent trade-off in 3D deep learning between the performance of voxel-based grids and the precision of point-based operations. Our work dismantles this compromise by introducing a computational system designed specifically for irregular point clouds.
By framing convolution on this data as an MVMR problem, we developed a dedicated GPU kernel that executes natively without the inefficient overheads. Our results confirm that sub-voxel accuracy could be retained without the performance penalties of prior methods, demonstrating that computational efficiency and geometric fidelity can be achieved in unison, enabling powerful, faithful geometric learning.

\subsubsection*{Acknowledgments}
This work was supported by Ant Group Research Intern Program. And we express our sincere gratitude to Jiaheng Li from Peking University for his assistance in setting up the Ant Group infrastructure used in the comparison experiments.